# A SYSTEM FOR PREDICTING SUBCELLULAR LOCALIZATION OF YEAST GENOME USING NEURAL NETWORK

Sabu M. Thampi[#], K. Chandra Sekaran[$]

\# Dept. of Computer Science & Engineering, L.B.S College of Engineering, Kasaragod
E-mail: smtlbs@yahoo.co.in

$ Department of Computer Engineering, National Institute of Technology Karnataka
E-mail: kch@nitk.ac.in

Abstract: The subcellular location of a protein can provide valuable information about its function. With the rapid increase of sequenced genomic data, the need for an automated and accurate tool to predict subcellular localization becomes increasingly important. Many efforts have been made to predict protein subcellular localization. This paper aims to merge the artificial neural networks and bioinformatics to predict the location of protein in yeast genome. We introduce a new subcellular prediction method based on a backpropagation neural network. The results show that the prediction within an error limit of 5– 10% can be achieved with the system.

# INTRODUCTION

A protein is a very large biological molecule composed of a chain of smaller molecules called amino acids. Thousands of different proteins are present in a cell, the synthesis of each type of protein being directed by a different gene. Proteins make up much of the cellular structure. Proteomics is concerned with qualitative and quantitative studies of gene expression at the level of the functional proteins themselves. Proteomics includes the identification, characterization and quantization of the entire complement of proteins in cells with a view to understanding their function in relation to the life of the cell.

An area of protein characterization that it considered particularly useful in the post-genomics era is the study of protein localization. In order to function properly, proteins must be transported to various localization sites within a particular cell. Description of protein localization provides information about each protein that is complementary to the protein sequence and structure data. The ability to identify known proteins with similar sequence and similar localization is becoming increasingly important, as we need structural, functional and localization information to accompany the raw sequences. The first approach for predicting the localization sites of proteins from their amino acid sequences was an expert system developed by Nakai and Kanehisa [6, 7]. Later, expert identified features were combined with a probabilistic model, which could learn its parameters from a set of training data [5]. Better prediction accuracy has been achieved by using standard classification algorithms such as K nearest neighbors (KNN), the binary decision tree, neural network and a nave Bayesian classifier.

This paper proposes a neural network-based approach for the subcellular localization of proteins. The paper is organized as follows. First we briefly describe the databases. Next we introduce various features. Next section demonstrates the system. This is followed by results and discussion section. Conclusion is given finally.

# DATABASES

The required database for localization is collected from the website http://bioinfo.mbb.yale.edu/genome/localize. There are four databases: Localized-465, Localized-704, Localized-1342, and Localized-2013. We have selected the database Localized-1342 because this dataset gives the best balance between overall quality and the number of proteins and it largely avoids the circular validation problem.

The Localized-1342 set is divided into seven subsets, each containing ~190 proteins. The proteins in each subset are selected randomly. Each protein belongs to only a single subset, and there are no duplicated proteins in any subset. The proteins in the Localized-1342 dataset are mainly associated with 12 subcellular compartments (Table 1) [1]. However, many of these compartments contain only a very small number of proteins, greatly skewing the statistics. Hence, the 12 compartments are collapsed into five new "generalized" compartments. The compartments are the nucleus (N), mitochondria (M), cytoplasm (C), membrane (T for transmembrane), and secretory pathway (E for endoplasmic reticulum or extracellular).

State vector database contains the probability state vector, which has 5 components, each one giving the probability that a protein can be found in the corresponding subcellular compartment. An example for the state vector of the protein, YAL001C is shown in Table 2. Here the first column indicates the name of the protein; second column indicates the actual compartment into which the protein is localized. Third column indicates the subset to which the protein belongs. The next five columns show the initial probability of the protein to be in the five locations stated earlier i.e.; C, N, M, T, and E. (In that dataset the values are given after multiplying by 1000).In the above example the probability of the protein to be in the location C is 0.001 and that in N is 0.997 and that in M is 0.0 and so on. In the feature vector dataset, for each vector there are five rows in the feature vector dataset, each one corresponding to C, N, M, T and E respectively. The value in each row shows the number of proteins in that compartment that possess the feature. Each component of the feature vector equals the fraction of the total number of proteins in that compartment possessing that feature.

Table 1. Collapsed Compartments

| Collapsed compartment | Compartments used for collapsing | |
|---|---|---|
| C | Cytoplasm | Cytosolic proteins (not in any organelles or membranes or cytoskeleton) |
| M | Mitochondria | Mitochondrial proteins |
| N | Nucleus | Nuclear proteins |
| T | Membrane Plasma membrane | Integral transmembrane proteins (in the cell membrane, the plasma membrane, or the membranes of various compartments such as mitochondria, nucleus, Golgi) |
| E | Endoplasmic reticulum(ER) Golgi apparatus, Peroxisome, vacuole, vesicle Extracellular | Proteins involved in the secretory pathway and those in small organelles |

Table 2. State vector of YAL001C

| scid_ | loc1 | subset | C | N | M | T | E |
|---|---|---|---|---|---|---|---|
| YAL001C | N | 5 | 1 | 997 | 0 | 2 | 0 |

## 2. FEATURES

Table 3 lists the features that are used in the system [1]. A total of 30 features are used. The features are first divided into three categories depending on the information they were derived from: (i) motifs (16 features); (ii) overall-sequence (four features); and (iii) whole-genome (ten features). The 30 features are subdivided into three groups, depending on how much they contributed to the overall predictive strength of our system: the ten most important features, nine other included features and 11 redundant features. For all the results reported here, the 11 redundant features are excluded and based our system on the best 19 features. For each feature, the proteins are divided into a specific number of bins according to the different values. Each bin can be used as a separate feature, and created a separate feature vector for each bin. The state vector of a protein is updated with feature vector of the bin it belongs to.

## DEMONSTRATING THE PROPOSED SYSTEM

Five prior state vectors and five feature vectors are used. There are 10 nodes in the input layer. The output layer consists of five nodes since we need the probability of the protein to be in the five compartments C, N, M, T and E. The hidden layers are selected by trial and error. The data is normalized so that it lies between 0 and 1. If the probability for a location in a feature vector is 0, or if a component of the state vector becomes 0 (after normalizing) while updating, the probability for the corresponding location in the resultant state vector will always be equal to 0. A pseudo-count of 0.0001 is added to all state vector components that are equal to 0, each time the state vector is updated. The transfer function used is sigmoid.

The network is trained using the first six subsets of the localized-1342 dataset whose compartment of localization is previously known. During cross-validation it is observed that the addition of some features with redundant information decrease the overall prediction accuracy. Hence we have selected only 13 features which raise the prediction accuracy. Table 3 summarizes the degree to which each feature raised or lowered the prediction accuracy when it is added to or subtracted from our system. This gives a rough measure of the "information content" of the feature.

In the training phase we took each protein one by one and applied 13 features one at a time. The desired output is the updated state vector after applying that particular feature. The actual output is the updated state vector obtained after applying that feature. The first feature is applied to the prior state vector of the protein obtained from the state vector dataset. Let P be a protein having state vector S(C, N, M, T, E) and X be the first feature applied to P. The inputs and output relationship is as shown in fig. 1. From the second feature onwards, the features are applied on the updated state vector obtained from applying the previous feature. Let Y be the next feature applied after X and let U(C, N, M, T and E) be the updated state vector obtained from applying feature X for protein P. The details are shown in fig. 2.

This process is repeated for all the 13 features. The updated state vector after the application of the 13 features will give us the actual location of the protein. The component of the state vector with the highest value will be the actual location of the protein. For example, if the updated state vector is (0.04, 0.07, 0.8, 0.03, 0.06), the protein is localized to the compartment M i.e. Mitochondria. At first the difference between the desired output and the actual output is huge and hence the mean square error is also large. Slowly the error comes down as the number of iterations through the proteins is increased. The network is trained repeatedly by changing the learning rate and momentum until we get converged.

Once the network is trained completely, the predicted location is matching with the actual location of the protein specified in the dataset. The demonstrated system has predicted the location of the proteins whose actual location is not known accurately. The details of a few of these proteins are listed in table 4.

**RESULTS AND DISCUSSION**

The proposed protein localization prediction system was implemented using C language. To make the software user friendly, we have provided a graphical user interface in Visual Basic. A few screen shots are shown in figures 3, 4, and 5. First user selects a protein. The result will be displayed in a bar chart form. The results show that the prediction within an error limit of 5–10% can be achieved with this method. This performance is higher than the Bayesian approach. Bayesian approach is able to predict only up to 75% accuracy on individual protein predictions.

Table 3. Features

| A. Brief description Feature | Type | Subtype | % Change | Status | Bin |
|---|---|---|---|---|---|
| MIT1 | Motif | Signal | 5.1 | Important | 2 |
| GLYC | Motif | Signal | 1.2 | Important | 1 |
| SIGNALP | Motif | Signal | 1.0 | Important | 2 |
| SIG1 | Motif | Signal | 0.7 | Important | 2 |
| NUC1 | Motif | Signal | 0.6 | Important | 6 |
| PI | Overall sequence | isoelectric point | 1.3 | Important | 10 |
| TMS1 | Overall sequence | Transmembrane helix | .9 | Important | 5 |
| MAYOUNG | Whole-genome | Absolute expr. (GeneChip) | 3.6 | Important | 10 |
| KNOCKOUT | Whole-genome | Knockout mutation | 1.8 | Important | 2 |
| MRDIASD | Whole-genome | Expr. fluctuation (Diauxic shift) | 1.4 | Important | 10 |
| PLMNEV1 | Motif | Signal | 0.3 | Included | 2 |
| FARN | Motif | Signal | 0.3 | Included | 2 |
| GGSI | Motif | Signal | 0.3 | Included | 2 |
| MIT2 | Motif | Signal | 0.2 | Included | 2 |
| HDEL | Motif | Signal | 0.1 | Included | 3 |
| NUC2 | Motif | Signal | 0.1 | Included | 6 |
| POX1 | Motif | Signal | 0.1 | Included | 2 |
| MRCYELU | Whole-genome | Expr. fluctuation (Cellcycle) | 0.4 | Included | 10 |
| MRCYCSD | Whole-genome | Expr. fluctuation (Cellcycle) | 0.2 | Included | 10 |
| COILDCO | Motif | Coiledcoils | -0.1 | Redundant | 2 |
| CKIISITE | Motif | Kinase target site | -0.1 | Redundant | 2 |
| CDC28SITE | Motif | Kinase target site | -0.3 | Redundant | 4 |
| PKASITE | Motif | Kinase target site | -0.5 | Redundant | 5 |
| ROSTALL | Overall sequence | Surface residue composition | -0.8 | Redundant | 9 |
| LENGTH | Overall sequence | Protein length | -1.6 | Redundant | 10 |
| MASAGEL | Whole-genome | Absolute expr (SAGE) | -0.3 | Redundant | 10 |
| MRCYC15 | Whole-genome | Expr. fluctuation (Cellcycle) | -0.4 | Redundant | 10 |
| MRCYC28 | Whole-genome | Expr. fluctuation (Cellcycle) | -0.6 | Redundant | 10 |
| MASAGEG | Whole-genome | Absolute expr | -0.7 | Redundant | 10 |
| MASAGES | Whole-genome | Absolute expr | -0.9 | Redundant | 10 |

Table 4: Proteins with Unknown Localizations

| scid_ | bc1 | subset | C | N | M | T | E |
|---|---|---|---|---|---|---|---|
| YAL002W | ? | 0 | 1 | 698 | 145 | 0 | 156 |
| YAL003W | ? | 0 | 970 | 27 | 0 | 0 | 3 |
| YAL004W | ? | 0 | 827 | 130 | 5 | 0 | 38 |
| YAL005C | ? | 0 | 529 | 262 | 1 | 0 | 208 |
| YAL008W | ? | 0 | 90 | 881 | 22 | 0 | 7 |
| YAL009W | ? | 0 | 19 | 980 | 0 | 0 | 1 |
| YAL010C | ? | 0 | 327 | 640 | 28 | 0 | 6 |
| YAL011W | ? | 0 | 0 | 1000 | 0 | 0 | 0 |
| YAL014C | ? | 0 | 196 | 755 | 38 | 0 | 11 |
| YAL015C | ? | 0 | 98 | 567 | 334 | 0 | 2 |
| YAL017W | ? | 0 | 458 | 527 | 4 | 0 | 11 |
| YAL018C | ? | 0 | 956 | 10 | 1 | 0 | 33 |
| YAL019W | ? | 0 | 0 | 1000 | 0 | 0 | 0 |
| YAL020C | ? | 0 | 633 | 312 | 50 | 0 | 5 |
| YAL021C | ? | 0 | 58 | 876 | 65 | 1 | 1 |
| YAL022C | ? | 0 | 482 | 37 | 2 | 0 | 480 |

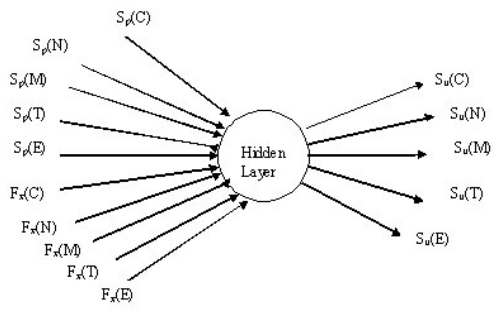 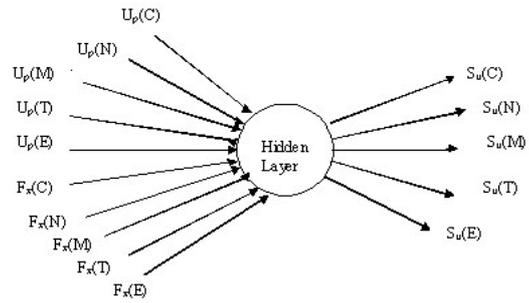

Fig. 1. Neural network output after applying state vector S and feature P

Fig 2. Neural network output after applying updated state vector U and feature Y

## CONCLUSION

An automatic, reliable and efficient prediction system for protein subcellular localization is needed for large-scale genome analysis. In this paper, a new method for protein subcellular localization prediction is presented. The network was trained and tested with data for predicting protein localization in yeast genome. The system is able to predict the location of the proteins whose actual location is unknown. The demonstrated method provides superior prediction performance as compared to Bayesian method.

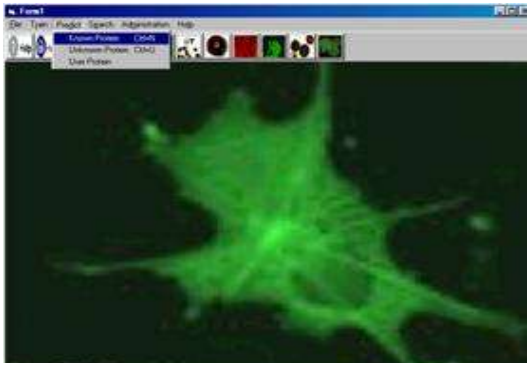 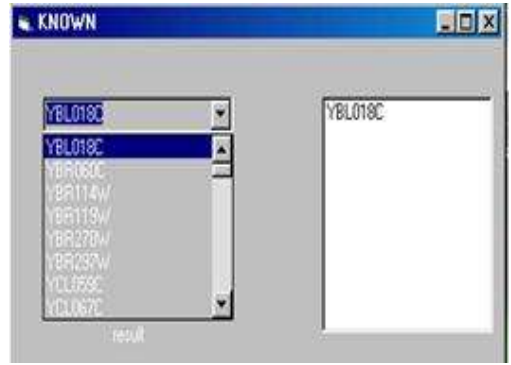

Fig 3. User Interface

Fig 4. User Selects a Protein

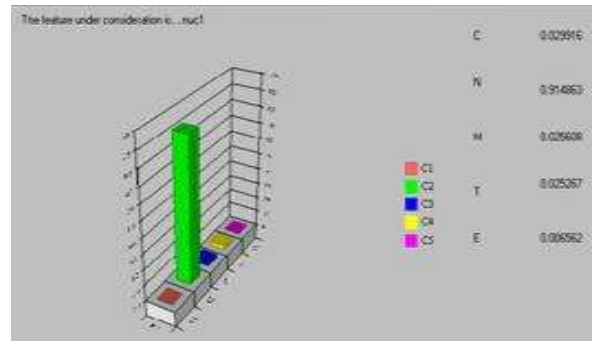

Fig 5. Result in a bar chart form